%% file: main.tex
\documentclass[10pt,twocolumn,letterpaper]{article}

\usepackage{iccv}              % To produce the CAMERA-READY version
\usepackage[accsupp]{axessibility}  % Improves PDF readability for those with disabilities.
\input{0_macros}

\definecolor{iccvblue}{rgb}{0.21,0.49,0.74}
\usepackage[pagebackref,breaklinks,colorlinks,allcolors=iccvblue]{hyperref}

\def\confName{ICCV}
\def\confYear{2025}

\title{Towards Efficient General Feature Prediction in Masked Skeleton Modeling}
\author{Shengkai Sun\textsuperscript{1}, Zefan Zhang\textsuperscript{2}, Jianfeng Dong\textsuperscript{3}, Zhiyong Cheng\textsuperscript{1}\thanks{Corresponding author: jason.zy.cheng@gmail.com} , Xiaojun Chang\textsuperscript{4}, Meng Wang\textsuperscript{1}\\
\textsuperscript{1}Hefei University of Technology \ \  \textsuperscript{2}Jilin University\\
\textsuperscript{3}Zhejiang Gongshang University \ \ \textsuperscript{4}University of Science and Technology of China\\
}

\begin{document}
\maketitle
\input{1_abstract}    
\input{2_intro}

\input{3_related_work}
\input{4_method}

\input{5_experiment}
\input{6_conclusion}
\section*{Acknowledgments}
This work was supported by the Pioneer and Leading Goose R\&D Program of Zhejiang (No. 2024C01110), National Natural Science Foundation of China (No. 72188101, 62020106007, and 62472385).
The computation is completed on the HPC Platform of Hefei University of Technology.
{
    \small
    \bibliographystyle{ieeenat_fullname}
    \bibliography{main}
}

\end{document}

%% file: 0_macros.tex
\usepackage{colortbl}
\usepackage{multirow}

\definecolor{lightcyan}{rgb}{0.88, 1.0, 1.0}

%% file: 1_abstract.tex
\begin{abstract}
Recent advances in the masked autoencoder (MAE) paradigm have significantly propelled self-supervised skeleton-based action recognition. However, most existing approaches limit reconstruction targets to raw joint coordinates or their simple variants, resulting in computational redundancy and limited semantic representation. To address this, we propose a novel \textbf{G}eneral \textbf{F}eature \textbf{P}rediction framework (\textbf{GFP}) for efficient mask skeleton modeling. Our key innovation is replacing conventional low-level reconstruction with high-level feature prediction that spans from local motion patterns to global semantic representations. Specifically, we introduce a collaborative learning framework where a lightweight target generation network dynamically produces diversified supervision signals across spatial-temporal hierarchies, avoiding reliance on pre-computed offline features. The framework incorporates constrained optimization to ensure feature diversity while preventing model collapse. Experiments on NTU RGB+D 60, NTU RGB+D 120 and PKU-MMD demonstrate the benefits of our approach: Computational efficiency (with 6.2$\times$ faster training than standard masked skeleton modeling methods) and superior representation quality, achieving state-of-the-art performance in various downstream tasks.

\end{abstract}

%% file: 2_intro.tex
\section{Introduction}
\label{sec:intro}

\begin{figure}[tb!]
\centering
\includegraphics[width=0.8\columnwidth]{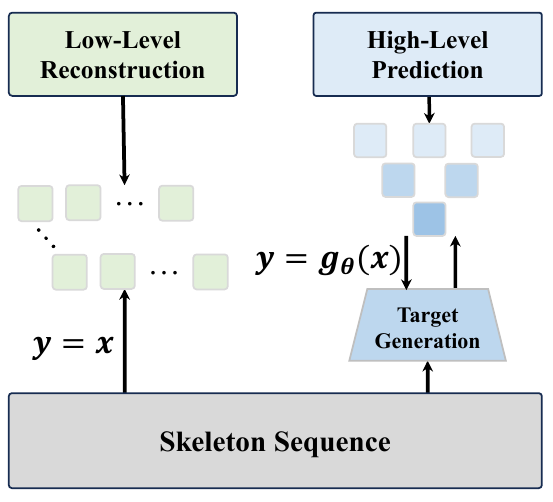}

\caption{Frameworks for general feature prediction (GFP). Instead of low-level targets, GFP predicts the hierarchical abstract representations of input skeleton sequence, embedded by a target generation network, effectively learning more informative high-level features for the downstream tasks. This paradigm achieves an optimal balance between model performance and computational efficiency. 
}\label{fig:intro_fig}
\end{figure}

Understanding human action \cite{ji20123d,feichtenhofer2019slowfast,zhao2017temporal} has long been a fundamental research task in computer vision, demonstrating significant application value in human-computer interaction, intelligent surveillance, and autonomous driving, \etc.
In contrast to conventional RGB video representations \cite{soomro2012ucf101, kuehne2011hmdb} that densely encode visual information, skeletal representations \cite{shahroudy2016ntu,liu2019ntu} capture body postures as topological joint sets, employing sparse coordinate sequences to characterize human actions instead of raw pixel values.
Owing to the compact data structure and explicit motion representation, skeleton-based action understanding \cite{yan2018spatial,shi2019two,shi2019skeleton,cheng2020skeleton} methods not only significantly reduce computational costs, but also effectively mitigate noise from dynamic backgrounds and lighting changes. However, despite these advantages, early approaches based on fully supervised learning faced notable limitations due to their heavy reliance on manually labeled skeleton data, which requires labor-intensive and costly annotation processes. To overcome this, the research community has increasingly focused on self-supervised learning \cite{zheng2018unsupervised,dong2023hierarchical,sun2023unified}, designing pretext tasks that can automatically learn spatiotemporal patterns in skeleton data without manual labels. The resulting representations can then be efficiently adapted to downstream tasks with a few labeled examples.

Self-supervised skeleton-based action learning evolved from early handcrafted methods (\eg, colorization, temporal segment rearrangement) to instance discrimination. Contrastive learning \cite{xu2024deep}, a prominent instance discrimination approach, gained dominance due to its simplicity and efficiency. However, current research reveals some limitations in contrastive learning: excessive dependence on data augmentation strategies and persistent false negative sample issues. Recently, transformers have achieved remarkable success in visual tasks \cite{dosovitskiy2020image, wang2024eulermormer,zhao2022tuber}, and transformer-based masked image modeling has emerged as a novel generative pretraining paradigm, offering fresh insights for self-supervised skeleton action learning.

The masked skeleton modeling methods follow the classic MAE framework by randomly masking some joints in the input skeleton sequence. Its encoder processes only the visible joints, while the decoder reconstructs missing joint coordinates from latent representations and learnable mask tokens. However, due to the intrinsic differences between skeleton and image data, skeleton MAE-like methods significantly differ in implementation. These methods employ a 90\% masking ratio to address temporal redundancy and adopt deeper decoders (instead of MAE's lightweight design) to handle complex spatiotemporal dependencies. Despite their impressive performance, the current design and learning objectives have clear shortcomings: (1) \textbf{Heavy decoder computation}: Due to the reconstruction target involving many joints and the transformer being used as the decoder, excessively long decoding sequences significantly slow down training; (2) \textbf{Lacking semantic guidance}: The low-level target reconstruction objective lacks explicit supervision from higher-level spatiotemporal features like motion semantics, which creates a semantic gap between pretraining and downstream tasks. These limitations motivate our exploration of new masked skeleton modeling paradigms that achieve a balance between representation quality and computational efficiency.

In this paper, we propose a novel \textbf{G}eneral \textbf{F}eature \textbf{P}rediction framework (\textbf{GFP}) for efficient mask skeleton modeling. As shown in \Cref{fig:intro_fig}, unlike methods targeting numerous redundant low-level features, GFP focuses on a smaller set of hierarchical high-level features, learning through progressive semantic prediction. Specifically, during decoding, it inputs elements of varying spatiotemporal granularity to predict from local motion patterns (\eg, a joint's movement over 10 frames) to global action semantics. This approach offers two advantages: First, the significantly reduced number of targets cuts computational costs in decoding and optimization. Second, it balances local details and global context, capturing both micro-motion changes and overall action structures, thereby substantially improving feature quality. 
To obtain these high-level semantic features as learning targets, we designed a lightweight target generation network. This network processes elements with varying spatiotemporal granularities, using a multiple feature extraction module to capture both local details and global context, providing hierarchical supervision for the decoder. This supervisory mechanism operates bidirectionally: The decoder infers high-level semantic features from unmasked joint features for learning, while the target generation network uses the decoder's output as an optimization target, guiding feature representation through a consistency learning manner. However, this design alone is insufficient, as joint optimization can lead to trivial solutions where both networks output constant values regardless of the input. Inspired by the recent information maximization representation learning method \cite{bardes2022vicreg}, we introduced a constraint mechanism to ensure meaningful feature outputs, enabling iterative optimization that converges to a semantically consistent stable state. Ultimately, we developed a flexible mask skeleton modeling framework that significantly accelerates training, enabling the model to learn more discriminative and robust feature representations, yielding superior performance across various downstream tasks.

Our contributions are summarized as follows:
\begin{itemize}
    \item We extend conventional mask skeleton modeling from low-level reconstruction to general feature prediction, improving decoding efficiency while introducing local and global semantic guidance.
    \item We design a lightweight target generation network that provides online supervision for the decoder. By introducing an information maximization constraint, we enable collaborative learning between networks, achieving a non-trivial self-supervised skeleton action representation learning scheme.
    \item Extensive experiments across three datasets demonstrate the effectiveness of our approach. The results show that our framework is highly efficient, learns richer semantic information, and outperforms previous methods on multiple downstream tasks.
\end{itemize}

%% file: 3_related_work.tex
\section{Related Work}
\label{sec:related_work}
\subsection{Self-supervised Skeleton-based Action Representation Learning}
Self-supervised skeleton-based action representation learning aims to learn discriminative features without manual annotation. The core challenge lies in designing meaningful supervisory signals to guide the model in capturing action semantics.

Early approaches to self-supervised skeleton action representation learning predominantly relied on handcrafted pretext tasks within encoder-decoder frameworks. Reconstruction-based methods formed the cornerstone: \cite{zheng2018unsupervised} reconstructed corrupted skeletons, while \cite{kundu2019unsupervised} and \cite{su2020predict} employed autoencoders for full-sequence reconstruction. \cite{nie2020unsupervised} pioneered feature disentanglement into pose/view components before reconstruction. Subsequent innovations leveraged spatio-temporal relationships—\cite{yang2021skeleton} predicted joint coloring which is pre-defined by spatial-temporal order, \cite{cheng2021hierarchical} forecasted inter-frame motions, and \cite{kim2022global} estimated multi-scale pose displacements. \cite{yang2024view} advanced view invariance through motion retargeting-driven reconstruction.

The paradigm shifted with the advent of contrastive representation learning. \cite{rao2021augmented} established the baseline by contrasting augmented skeleton pairs via dual encoders. \cite{thoker2021skeleton} enhanced this with skeleton-specific augmentations and cross-architectural contrast. Later works diversified positive pair generation: \cite{li20213d, zhang2022contrastive} mined contextual positives, \cite{shah2023halp} synthesized latent-space positives.
Multi-modal fusion emerged through \cite{mao2022cmd} and \cite{sun2023unified}. \cite{Hua2023SkeAttnCLR,lin2023actionlet} introduced guidance mechanisms focusing on action-critical regions to enhance semantic feature learning. \cite{zhang2023prompted} pioneered dual-prompted pretraining combining masked reconstruction and feature contrast.

Building on the success of masked autoencoders, recent works \cite{wu2023skeletonmae,mao2023masked,abdelfattah2024s} have adapted this paradigm to skeleton-based action representation learning, demonstrating impressive results in capturing action semantics through partial observation and reconstruction.
\\
\\
\subsection{Masked Modeling for Skeleton-based Action}
\cite{wu2023skeletonmae} pioneers the adaptation of Masked Autoencoder to transformer-based skeleton-based action representation learning, employing direct reconstruction of masked joint coordinates as its pretext task. \cite{mao2023masked} extended SkeletonMAE to motion feature prediction, proposing a motion-aware masking strategy that prioritizes high movement intensity regions during input masking. This input masking forces the model to focus on action-critical patches, achieving substantial performance improvements over conventional random masking approaches. 
Despite these advancements, the reliance on low-level reconstruction targets inherently limits high-level semantic learning, creating a performance gap in tasks requiring holistic action representation. \cite{abdelfattah2024s} attempted to address this by using model-generated features as learning targets. However, their patch-level feature prediction necessitates a bulky EMA-updated encoder of massive computations for target generation.
While achieving accuracy gains, this approach incurs higher computational costs and much slower convergence than conventional methods.

By contrast, our method employs spatiotemporally high-level semantic prediction paired with a lightweight target generation network. This design achieves much faster training versus \cite{abdelfattah2024s} and higher downstream accuracy.

%% file: 4_method.tex
\section{Method}

\begin{figure*}[tb!]
\centering\includegraphics[width=2.0\columnwidth]{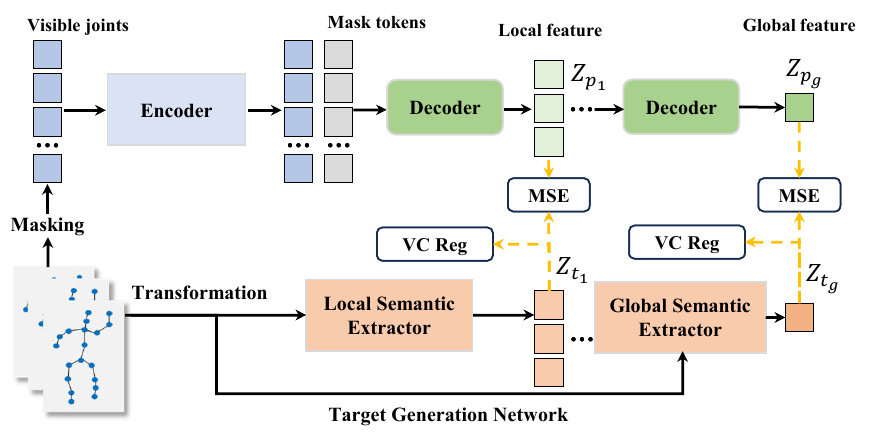} \vspace{-2mm}
\caption{
The framework of our proposed GFP. The GFP framework establishes a bidirectional learning paradigm between two core networks: an encoder-decoder architecture and a Target Generation Network (TGN). The encoder extracts features from partially visible skeleton joints, while the decoder progressively predicts hierarchical high-level features — spanning short-term motion patterns to global action semantics. Simultaneously, the TGN generates semantic targets by encoding transformed inputs and performs consistency learning. This mutual guidance is stabilized through variance-covariance regularization, an information-maximization constraint that ensures feature diversity. 
}
\label{fig:framework}
\end{figure*}
In this section, we first reformulate masked skeleton modeling by replacing its learning objective from conventional patch-level reconstruction to high-level semantic feature prediction. Next, we develop a hierarchical feature prediction framework through multi-granularity representation learning, jointly predicting local details and global semantics. Finally, we design a dedicated target generation network to provide high-quality supervision for prediction tasks, coupled with an information constraint mechanism ensuring stable model optimization.

\subsection{Overview}
The pipeline of skeleton masked autoencoders operates as follows: A predefined masking strategy randomly masks joints, training the network to reconstruct masked joints for representation learning. Specifically, given an input skeleton sequence ${X} \in \mathbb{R}^{T \times V \times C}$, where $T$ is the number of frames, $V$ is the number of joints, and $C$ is the number of channels, temporal patchify first transforms it into ${X_e} \in \mathbb{R}^{(T_e \times V) \times (l \times C)}$. This operation segments the temporal streams of every joint into non-overlapping $l$-frame patches based on consideration of computational efficiency, where $T_e=T/l$. The patches are embedded into feature vectors ${E} \in \mathbb{R}^{(T_e \times V) \times C_e}$ of $C_e$ dimension. Most patches are randomly masked, leaving only a small subset visible as input to the transformer encoder for feature extraction. The decoder subsequently predicts masked regions using sequence $E_N \in \mathbb{R}^{(T_e \times V) \times C_e}$ which combines extracted visible features with learnable mask tokens:
\begin{equation}\label{eq:mse}
      \mathcal{L}_{p} = \frac{1}{N}  \|f(E_N) - X_e\|_F^2,
\end{equation}
where $N=T_e \times V$, $f(E_N)$ is the prediction of decoder $f()$. Prediction targets in the current works extend beyond raw coordinates to include motion trajectories (\ie, frame differences \cite{mao2023masked}) or patch features from other networks \cite{abdelfattah2024s}. 

However, such low-level targets share critical limitations: high computational costs and inadequate high-level semantic capture, creating significant gaps for downstream tasks. This drives our extension of prediction targets to different high-level features through:
\begin{equation}\label{eq:mse}
      \mathcal{L}_{p} = \frac{1}{M}  \|f(E_N) - g(X)\|_F^2,
\end{equation}
where $M$ is the number of target features, $g$ is a network to extract various features from the input skeleton action as targets of the decoder, such as action-aware representative features. This high-level feature prediction enhances feature transferability for downstream tasks, particularly those demanding global semantic understanding like recognition and retrieval.

\subsection{High-level Feature Prediction}
To achieve the above learning objectives, we develop a hierarchical architectural instantiation that strategically balances local motion details with global contextual awareness. Our approach focuses on selecting learning objectives with richer semantic information across larger spatial-temporal scales, such as the motion patterns of specific nodes over 10-frame sequences. To handle numerous candidate features, we designed a progressive modeling strategy. It starts with short-term local joint patterns, gradually expands to full cycles, and finally integrates local features for holistic action prediction.

Formally, the decoder takes $E_N \in \mathbb{R}^{(T_e \times V) \times C_e}$ that combines unmasked joint features from the encoder with mask tokens as input. The pipeline first employs temporal average pooling with kernel size $t_1$, then processes the sequence through a transformer decoder to build the local motion prediction model, producing feature prediction $Z_{p_1} \in \mathbb{R}^{((T_e/t_1) \times V) \times C_p}$. Following the initial stage, our hierarchical architecture iteratively scales temporal receptive fields through cascaded processing blocks. For the second stage, temporal average pooling with another kernel size is applied to $Z_{p_1}$, followed by another transformer decoder that models mid-term dynamics, yielding $Z_{p_2} \in \mathbb{R}^{((T_e/t_2) \times V) \times C_p}$. This pyramidal reduction continues until capturing complete action cycles, where the temporal dimension collapses to a singleton, gaining different time granularity feature predictions $\{Z_{p_1}, Z_{p_2}, \ldots, Z_{p_k}\}$. The compressed representation $Z_{p_k} \in \mathbb{R}^{V \times C_p}$ then passes through feature aggregation, followed by an MLP for action-level semantic prediction $Z_{p_g}$. This multi-granularity fusion enables simultaneous preservation of fine-grained joint kinematics and high-level semantic context. Upon obtaining the hierarchical features, our encoder-decoder network learns by minimizing discrepancies between predicted features and target features across hierarchy levels. Given the target set $\{Z_{t_1}, Z_{t_2}, \ldots, Z_{t_k}, Z_{t_g}\}$ and $\mathcal{J} = {1,...,k,g}$ , the loss function is formulated as:  
\begin{equation}\label{eq:mse}
      \mathcal{L}_{p} = \frac{1}{M} \sum_{j \in \mathcal{J}} \|Z_{p_j} - Z_{t_j}\|_F^2.
\end{equation}
where $M$ is the number of all targets. All predicted features are dimensionally aligned to target features through an MLP projector, retaining the original notation for brevity.

It is worth noting that previous work predicting patch features from other networks can be viewed as a special case of our method when $t_1=1$ with only a single hierarchy level. However, such exclusive reliance on the finest-grained (yet most redundant) local information inevitably leads to significant computational overhead while struggling to capture global contextual patterns.  

\subsection{Target Generation}

\textbf{Semantic Extraction:} To obtain the aforementioned high-level semantic targets, we introduce a jointly optimized target generation network (\textbf{TGN}) composed of multiple lightweight MLPs, where each MLP specializes in producing hierarchical semantic features at specific granularity. Specifically, for each hierarchy level, we adopt simple spatiotemporal aggregation operations that process input signals within corresponding receptive fields: local-level semantic extractors operate on joints that merge across consecutive frames, while the global-level semantic extractor performs holistic feature extraction by integrating all spatiotemporal context. As an example, for the target feature $Z_{t_1}$, the corresponding input is  represented as $X_{t_1} \in \mathbb{R}^{((T_e/t_1) \times V) \times (t_1 \times l \times C)}$.

As for the input of TGN, feeding identical encoder input $\mathbf{X}_e$ to both networks amplifies bias towards frequently unmasked joints. We devise a simple transformation strategy to get $X_e = X_e[1:] - X_e[:-1]$. This geometrically grounded transformation preserves high-level motion semantics while avoiding bias. Our experiments demonstrate the remarkable effectiveness of this parameter-efficient design.

\textbf{Information Maximization:} Under the current framework where the decoder learns from TGN-generated targets while the TGN simultaneously adapts to the decoder's evolving feature representations, this co-learning scheme enables consistent high-level semantic alignment between both networks. However, such bidirectional dependency inevitably risks collapsing into trivial solutions where both networks output constant feature vectors independent of input signals, thereby losing all discriminative power. To prevent this degenerative collapse, we introduce a self-constrained feature space regularization mechanism inspired by VICReg \cite{bardes2022vicreg}, a joint embedding architecture that enforces information-maximization constraints.

\textit{Variance Regularization.} Taking the global target feature $Z_{t_g} \in \mathbb{R}^{B \times C_t}$ as an example, variance regularization enforces each feature dimension to maintain batch-wise variance above a predefined threshold $\gamma$, ensuring persistent diversity in the learned representations and preventing pathological collapse where disparate inputs are mapped to identical feature vectors. Formally, this is implemented through a variance regularization term:  
\begin{equation}
\mathcal{L}_{var} = \frac{1}{C_t}\sum_{i=1}^{C_t} \max\left(0, \gamma - \sqrt{ \mathrm{Var}(Z_{t_g}[:,i])}\right)
\end{equation}  
where $\mathrm{Var}(\cdot)$ computes empirical variance over the batch dimension $B$. Following \cite{bardes2022vicreg}, $\gamma$ is set to 1 in our experiments.

\textit{Covariance Regularization.} The covariance constraint drives the batch-wise covariance between every feature dimension pair toward zero, eliminating inter-dimensional correlations to prevent redundant information encoding across channels. This is formalized through a covariance minimization term:  
\begin{equation}
\mathcal{L}_{cov} = \frac{1}{C_t}\sum_{i \neq j} \left[\mathrm{Cov}(Z_{t_g})\right]_{i,j}^2
\end{equation}  
where $Cov(Z_{t_g})$ denotes covariance matrix of $Z_{t_g}$. 

\textbf{Optimization Objectives: }The final information-maximization constraints are applied to all target features across hierarchy levels through unified regularization:  
\begin{equation}
\mathcal{L}_{reg} = \sum_{j \in \mathcal{J}} \left( \alpha \mathcal{L}_{cov}{(Z_{t_j})} + \beta \mathcal{L}_{var}{(Z_{t_j})} \right), 
\end{equation}  
where $\alpha$ and $\beta$ are trade-off coefficients. The collaborative optimization objective for both networks combines the prediction loss with information-maximization constraints through weighted aggregation:
\begin{equation}
\mathcal{L}_{total} = \lambda \mathcal{L}_{pred} + \mathcal{L}_{reg} 
\end{equation}
where $\lambda$ denotes trade-off paramters, and ${L}_{pred}$ denotes the prediction loss computed over the entire batch, \ie, ${L}_{pred}=\frac{1}{B}\sum \mathcal{L}_{p}$. This unified objective drives the co-evolution of the decoder and TGN towards both precise hierarchical alignment and collapse-resistant feature extraction.

%% file: 5_experiment.tex
\section{Experiments}

\subsection{Datasets}
\input{tables/comparison}

\input{tables/ntu120_linear_eval}

\input{tables/knn}
This paper conducts experiments on three widely-used skeleton action benchmarks: NTU RGB+D 60 (NTU-60), NTU RGB+D 120 (NTU-120), and PKU-MMD II (PKU-II). Technical specifications and evaluation protocols for these datasets are summarized as follows.

\textbf{NTU RGB+D 60}~\cite{shahroudy2016ntu} features 60 action classes with 56,880 sequences captured from 40 participants using multi-view cameras. Two standard evaluation settings are adopted: 1) Cross-Subject (x-sub) divides data by participant IDs, using 20 subjects for training and 20 for testing; 2) Cross-View (x-view) employs camera views 2-3 for training and view 1 for testing.

\textbf{NTU RGB+D 120}~\cite{liu2019ntu} extends NTU-60 to 120 action classes with 114,480 sequences from 106 subjects across 32 capture setups. Its evaluation protocols include: 1) Cross-Subject (x-sub) split with 53 subjects each for training/testing; 2) Cross-Setup (x-setup) partitioning using even-numbered setups for training and odd-numbered for testing.

\textbf{PKU-MMD II}~\cite{liu2020benchmark} focuses on viewpoint-robust action analysis with 51 categories containing 6,952 sequences. The dataset's complexity stems from significant viewpoint diversity. Following established convention, we report results under the standard cross-subject split (5,339 training/1,613 test samples).

Consistent with recent literature~\cite{thoker2021skeleton,mao2022cmd}, all experiments on three datasets employ top-1 accuracy as the evaluation metric.

\subsection{Implementation Details}
\textbf{Model Architecture.}  
Follow \cite{mao2023masked, abdelfattah2024s}, our model utilizes a standard vision transformer backbone with learnable spatio-temporal position encoding. The encoder stacks $L_e=8$ identical layers, each containing multi-head attention (8 heads, 256-dim) and a feed-forward network (1024 hidden units). For pre-training, the decoder retains the same dimensions but simplifies the architecture: each local feature decoder uses one layer, while the global feature decoder adopts a 2-layer MLP with 2048-dimensional hidden units. The target generation network employs a 3-layer MLP (512 hidden units) for local semantic extraction and a 3-layer MLP (2048 hidden units) for global semantic extraction. All projectors in the decoder are 3-layer MLPs that project features to match the dimensions of the target features.
\\
\textbf{Data Augmentation.}  
For skeleton sequence processing, as in previous works \cite{mao2023masked, abdelfattah2024s}, we adopt temporal cropping with dynamic scaling by randomly selecting a continuous clip spanning 50\%-100\% of the original sequence length. All clips are resampled to frames via bilinear interpolation to ensure temporal consistency.
\\
\textbf{Pretraining Setup.}  
We employ the motion-aware masking strategy proposed in \cite{mao2023masked} and segment length $l$ to 4 in the pertaining. Optimization uses AdamW ($\beta_1=0.9$, $\beta_2=0.95$, weight decay 0.05) under a 400-epoch schedule. Learning rates ramp up linearly from 0 to 1e-3 during 20 warmup epochs, then decay to 5e-4 via cosine annealing. Hyperparameters $\lambda$, $\alpha$ and $\beta$ are set to 5, 5, 1, respectively. For the target features, we employ a four-level hierarchical feature corresponding to local features at $t_1=5$, $t_2=10$, $t_3=30$, and the global feature.

\subsection{Comparison to the State-of-the-art}
In this section, we compare our method with state-of-the-art approaches on two skeleton-based tasks: action recognition and action retrieval.

\textbf{Skeleton-based Action Recognition.} 
Following the standard practice from previous works \cite{thoker2021skeleton,mao2023masked, abdelfattah2024s}, we train a linear classifier on top of the frozen encoder pre-trained with our proposed method. First, we compare our method against prior MAE-like approaches \cite{wu2023skeletonmae,mao2023masked,abdelfattah2024s}, on the NTU-60 dataset. Beyond recognition performance, we quantify computational efficiency by measuring computational FLOPs per action sequence for encoder, decoder, and target generation network, full pre-training time under the x-sub protocol (identical hardware), and training speedup relative to SkeletonMAE \cite{wu2023skeletonmae}. As shown in \Cref{table:ntu60_linear_eval}, our method achieves significant performance gains while substantially accelerating training. This improvement stems from reducing the optimization targets from \textbf{750 low-level targets (N=750)} in prior works to \textbf{251 hierarchical high-level semantic targets (M=251)}. Notably, while S-JEPA also adopts learned features as objectives, its patch-level implementation necessitates redundant feature encoding. This design choice, coupled with extremely slow convergence (requiring 1200 epochs versus 400 epochs for other methods), severely degrades training speed compared to SkeletonMAE. These results demonstrate our method's balanced efficacy in feature quality and computational efficiency.

We further validate our approach on larger-scale NTU-120 and more challenging PKU-MMD II datasets, comparing against methods using alternative pretext tasks. As shown in \Cref{table:ntu120_linear_eval}, our method outperforms most existing approaches, confirming its broad effectiveness.

\begin{figure}[tb!]
\centering
\includegraphics[width=3.2in]{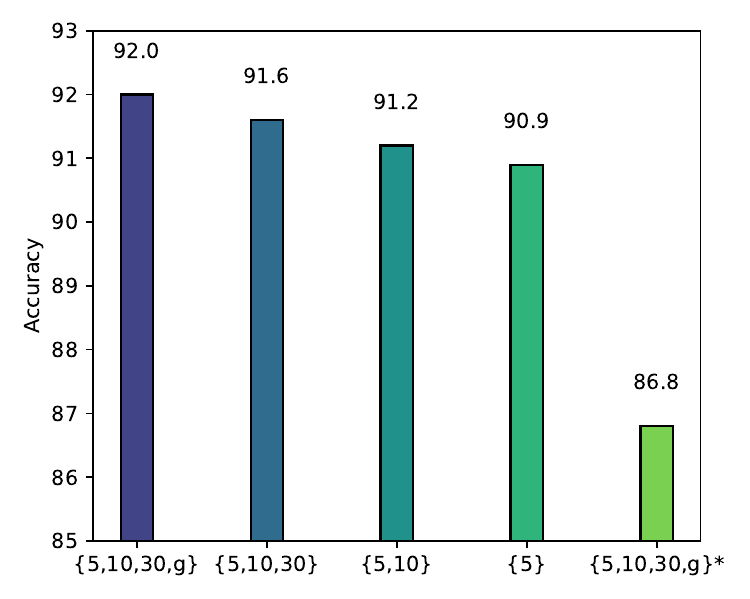}
\vspace{-4mm}
\caption{Impact of Different Targets. The {5,10,30,g} configurations represent local features from 5-, 10-, 30-frame temporal spans and global representations respectively, all derived from high-level representations generated by the TGN. All results are obtained under the NTU-60 x-view protocol. 
} \label{fig:targets}
\end{figure}

\textbf{Skeleton-based Action Retrieval.} 
In this experiment, the encoder operates in a direct feature extraction mode without requiring any training. Similar to video-to-video retrieval, we use cosine similarity in the feature space to retrieve the action samples for a given query. As detailed in \Cref{tab:knn}, conventional MAE-like methods exhibit poor performance on this task due to their focus on local reconstruction objectives without global semantic learning. In contrast, our method achieves substantial improvements (8.9\% on x-sub and 17.1\% on x-view protocols) over prior approaches like MAMP by incorporating high-level semantic objectives. Notably, our results remain competitive with contrastive learning methods \cite{dong2023hierarchical,sun2023unified} that explicitly optimize global representations, demonstrating the superiority of high-level semantic targets over low-level reconstruction objectives.

\subsection{Ablation Study}
All ablation studies were conducted on the NTU-60 dataset for skeleton-based action recognition.

\begin{figure}[tb!]
\centering
\includegraphics[width=3.2in]{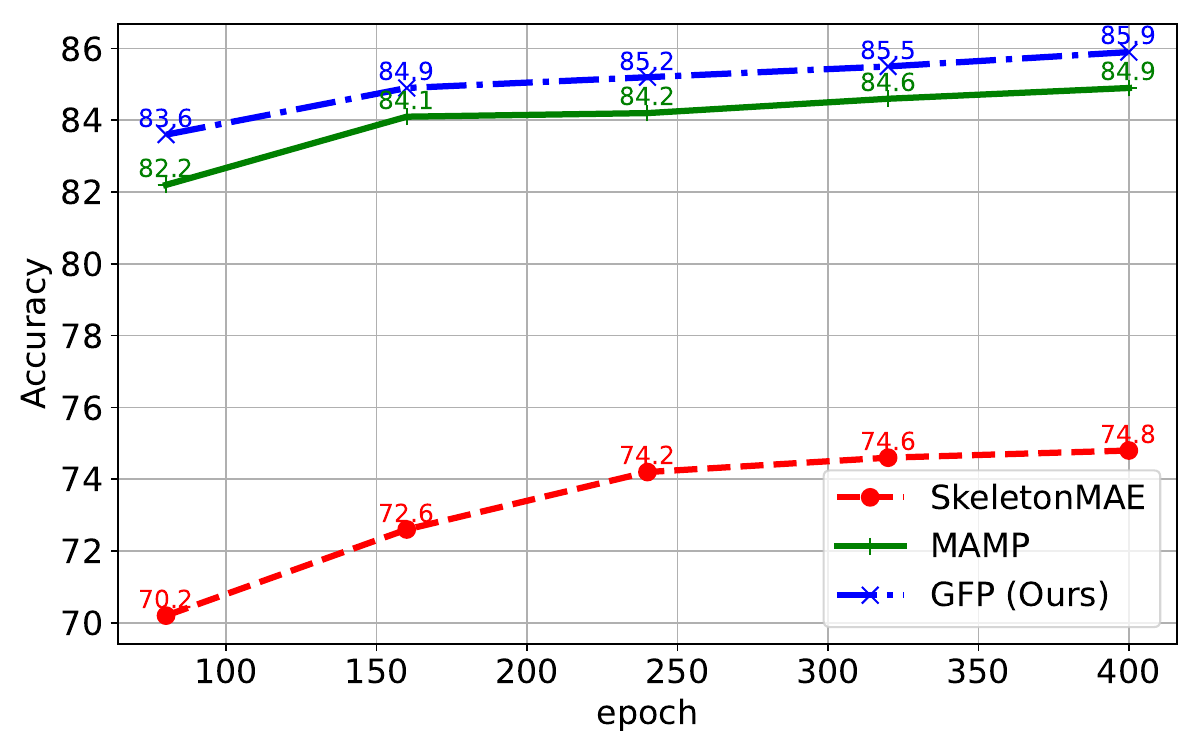}
\vspace{-4mm}
\caption{Performance of various methods with different targets. All results are obtained under the NTU-60 x-sub protocol. 
} \label{fig:epoch}
\end{figure}

\textbf{Prediction Targets: }Our method employs local joint features (5-, 10-, and 30-frame spans) and global semantic features as learning targets. To investigate their individual impacts, we progressively ablated these components. As shown in \Cref{fig:targets}, sequential removal of global and local semantic learning objectives caused performance degradation, demonstrating the complementary nature of hierarchical feature learning. Furthermore, we explored using low-level targets instead of learned representations by removing the TGN from our framework, which forces direct prediction vectors of all spatiotemporal information within targets (\eg, flattening input into a single vector for global target). The corresponding results (labeled \{5,10,30,g\}* in \Cref{fig:targets}) reveal significant performance drops, proving that high-level semantics cannot be effectively captured through direct prediction of low-level signals.
\input{tables/ablations}
\input{tables/semi_supervised}

\textbf{Input of Target Generation Network: }For the target network inputs, we evaluated three variants: (1) raw joint coordinates, (2) masked joint data, and (3) motion features derived via linear transformation of joints. As shown in \Cref{tab:sub1}, the motion features achieved optimal performance, validating our design choice. Notably, masked inputs hinder semantic feature extraction by TGN, aligning with observations in prior work \cite{abdelfattah2024s}. We evaluate the epoch-wise performance of different methods with varying learning objectives: SkeletonMAE learns to reconstruct raw joints, MAMP targets motion features (i.e., temporal difference of joint coordinates), and our method predicts high-level semantic representations. As shown in \Cref{fig:epoch}, our approach consistently outperforms prior works across all training epochs, demonstrating the superiority of hierarchical semantic learning over low-level reconstruction objectives.

\textbf{Architecture Design: } We conducted architectural experiments on the target generation network and projector. For TGN: (1)Shallow variant: 2-layer MLP;
(2)Wider variant: 4-layer MLP with doubled hidden dimensions. (3)Base: Standard 3-layer MLP. As shown in \Cref{tab:sub2}, the baseline 3-layer configuration achieves optimal performance, though all variants show marginal differences, demonstrating TGN's robustness to architectural variations. For the projector, we evaluated MLPs with 2-4 layers. The 3-layer configuration yields superior performance (\Cref{tab:sub3}).

\subsection{Extended Performance Validation}
To assess representation transferability, we conduct limited-label experiments on NTU-60 under two data settings: 1\%  and 5\% labeled training data. Following semi-supervised learning benchmarks~\cite{thoker2021skeleton}, we initialize the model with weights pre-trained through our model, then attach and fine-tune a classifier using available labels.

As shown in Table~\ref{tab:sota-semisupervised}, our approach establishes new state-of-the-art performance across both data scales. This significant performance gap confirms the effectiveness of our learned representations in label-efficient scenarios.

%% file: tables/comparison.tex
\begin{table*}[t]
\centering
{
\caption{\textbf{Skeleton-based action recognition.} Comparisons to the state-of-the-art MAE-like methods for skeleton-based action recognition downstream task on NTU-60. All pre-training runs were completed under identical hardware conditions using a single RTX 4090 GPU. Enc, Dec, and TGN denote the encoder, decoder, and target generation network, respectively.
}
\label{table:ntu60_linear_eval}
}

\begin{tabular}{@{}ll* {9}c l@{}}
\toprule
& \multirow{2}{*}{\textbf{Method}}  &
\multirow{2}{*}{\textbf{Target}}   &
\multirow{2}{*}{\textbf{Enc}}   &
\multirow{2}{*}{\textbf{Dec}}   &
\multirow{2}{*}{\textbf{TGN}}   &
\multirow{2}{*}{\textbf{Hours}}   &
\multirow{2}{*}{\textbf{Speedup}}   &
\multicolumn{2}{c}{\textbf{NTU-60}} &
\\
 \cmidrule(r){9-10}
&&&&&&&& x-sub & x-view  \\
\midrule 
&SkeletonMAE \cite{wu2023skeletonmae} & Joint & 1.97G & 17.70G & - & 20h27m & 1$\times$ & 74.8 & 77.7 \\
&MAMP \cite{mao2023masked} & Motion & 1.97G & 17.70G & - & 20h27m & 1$\times$ & 84.9 & 89.1 \\
&S-JEPA \cite{abdelfattah2024s} & Patch-level & 1.97G & 17.70G & 28.32G & 90h57m & 0.2$\times$ & 85.3 & 89.8 \\
\rowcolor{lightcyan}
&GFP (Ours) & Hierarchical & 1.97G & 1.57G & 0.64G & \textbf{3h14m} & \textbf{6.2$\times$} & \textbf{85.9} & \textbf{92.0} &\\

\bottomrule
\end{tabular}

\vspace{-0.4cm}
\end{table*}

%% file: tables/ntu120_linear_eval.tex
\begin{table}[t]
\centering
{
\caption{\textbf{Skeleton-based action recognition.} Comparisons to the state-of-the-art methods for skeleton-based action recognition downstream task on NTU-120 and PKU-MMD II. 
\textbf{Bold} and \underline{underline} formatting indicate the best and second-best methods in each group, respectively.
}
\label{table:ntu120_linear_eval}
}
\begin{tabular}{@{}ll* {4}c l@{}}
\toprule
& \multirow{2}{*}{\textbf{Method}}  &
\multicolumn{2}{c}{\textbf{NTU-120}} &
\multicolumn{1}{c}{\textbf{PKU-II}}
\\
\cmidrule(r){3-4} \cmidrule(r){5-5}
& & x-sub & x-setup & x-sub \\
\midrule 
&\textbf{\textit{other pretext task:}} &  \\
&P\&C \cite{su2020predict} & 42.7 & 41.7 & 25.5 \\
&ISC \cite{thoker2021skeleton} & 67.1 & 67.9 & 36.0 \\
&HaLP\cite{shah2023halp}  & 71.1 & 72.2 & 43.5 \\
&HiCo\cite{dong2023hierarchical}  & 72.8 & 74.1 & 49.4 \\
&CMD \cite{mao2022cmd} & 70.3 & 71.5 & 43.0 \\
&UmURL\cite{sun2023unified}  & \underline{73.5} & \underline{74.3} & \underline{52.1} \\
&USDRL \cite{weng2024usdrl} & \textbf{76.6} & \textbf{78.1} & \textbf{54.4} \\
\midrule
&\textbf{\textit{masked modeling:}} &  \\
&SkeletonMAE\cite{yang2023self} & 72.5 & 73.5 & 36.1 \\
&MAMP \cite{mao2023masked} & 78.6 & 79.1 & \underline{53.8} \\
&S-JEPA \cite{abdelfattah2024s} & \textbf{79.6} & \underline{79.9} & 53.5 \\
\rowcolor{lightcyan}
&GFP (Ours) & \underline{79.1} & \textbf{80.3} & \textbf{56.2} \\

\bottomrule
\end{tabular}

\vspace{-0.4cm}
\end{table}

%% file: tables/knn.tex
\begin{table} [tb!]
\renewcommand{\arraystretch}{1.2}
\caption{Comparisons to the state-of-the-art methods for skeleton-based action retrieval. \textbf{Bold} indicates the best methods in each group.
}
\label{tab:knn}
\centering 
\scalebox{0.9}{
\begin{tabular}{@{}ll* {3}c @{}}
\toprule
&\multirow{2}{*}{\textbf{Method}}   & 
\multicolumn{2}{c}{\textbf{NTU-60}} & \\
\cmidrule(r){3-4} 
&& x-sub & x-view  \\
\cmidrule{1-4}
&\textbf{\textit{other pretext task:}} &  \\
&LongT GAN\cite{zheng2018unsupervised} & 39.1 & 48.1 & \\
&P\&C\cite{su2020predict} & 50.7 & 76.3  \\\
&ISC\cite{thoker2021skeleton} & 62.5 & 82.6  & \\
&HaLP\cite{shah2023halp} & 65.8 & 83.6  \\
% &KTCL\cite{10539295}(TMM'24) & 67.1 & 84.7  \\
&HiCo\cite{dong2023hierarchical} & 68.3 & 84.8  \\
&SkeAttnCLR\cite{Hua2023SkeAttnCLR} & 69.4 & 76.8 &  \\
&UmURL\cite{sun2023unified} & \textbf{71.3} & \textbf{88.3}  \\
\cmidrule{1-4}
&\textbf{\textit{masked modeling:}} &  \\
& MAMP \cite{mao2023masked} & 62.0 & 70.0 & \\
\rowcolor{lightcyan}
&GFP (Ours) & \textbf{70.9} & \textbf{87.1}     \\
\bottomrule
\end{tabular}
}
\end{table}

%% file: tables/ablations.tex
\begin{table*}[ht]
  \centering
  \begin{minipage}[t]{0.3\textwidth}
    \centering
    \caption{Ablation study on the input of target generation network.
    }
    \label{tab:sub1}
    \begin{tabular}{@{}l*{3}c @{}}
    \toprule
    \multirow{2}{*}{\textbf{Input}}   & 
    \multicolumn{2}{c}{\textbf{NTU 60}} &\\
    \cmidrule{2-3} 
    & \multicolumn{1}{c}{\textbf{x-sub}} &
    \multicolumn{1}{c}{\textbf{x-view}} &\\
    \cmidrule{2-3} 
    \cmidrule{1-4}
    joint & 85.0 & 90.9 \\
    masked joint & 84.2 & 90.3 \\
    motion & 85.9 & 92.0 \\
    \bottomrule
    \end{tabular}
  \end{minipage}
  \hfill
  \begin{minipage}[t]{0.3\textwidth}
    \centering
    \caption{Ablation study on the architecture of target generation network.
    }
    \label{tab:sub2}
    \begin{tabular}{@{}l*{3}c @{}}
    \toprule
    \multirow{2}{*}{\textbf{Arch.}}   & 
    \multicolumn{2}{c}{\textbf{NTU 60}} &\\
    \cmidrule{2-3} 
    & \multicolumn{1}{c}{\textbf{x-sub}} &
    \multicolumn{1}{c}{\textbf{x-view}} &\\
    \cmidrule{2-3} 
    \cmidrule{1-4}
    Base & 85.9 & 92.0 \\
    Shadow & 85.8 & 91.8 \\
    Wider & 85.8 & 91.3 \\
    \bottomrule
    \end{tabular}
  \end{minipage}
  \hfill
  \begin{minipage}[t]{0.3\textwidth}
    \centering
    \caption{Ablation study on different number of layers used in projector.
    }
    \label{tab:sub3}
    \begin{tabular}{@{}l*{3}c @{}}
    \toprule
    \multirow{2}{*}{\textbf{Layers}}   & 
    \multicolumn{2}{c}{\textbf{NTU 60}} &\\
    \cmidrule{2-3} 
    & \multicolumn{1}{c}{\textbf{x-sub}} &
    \multicolumn{1}{c}{\textbf{x-view}} &\\
    \cmidrule{2-3} 
    \cmidrule{1-4}
    2 & 85.5 & 91.3 \\
    3 & 85.9 & 92.0\\
    4 & 85.6 & 91.6\\
    \bottomrule
    \end{tabular}
  \end{minipage}
\vspace{-0.4cm}
\end{table*}

%% file: tables/semi_supervised.tex
\begin{table} [tb!]
\renewcommand{\arraystretch}{1.2}
\caption{Comparisons to the state-of-the-art methods with semi-supervised learning on NTU-60 dataset.
}
\label{tab:sota-semisupervised}
%\vspace{-3mm}
%
\centering 
\scalebox{1.0}{
\begin{tabular}{@{}l*{5}c @{}}
\toprule
\multirow{2}{*}{\textbf{Method}}   &
\multicolumn{2}{c}{\textbf{x-sub}} & \multicolumn{2}{c}{\textbf{x-view}} \\

\cmidrule(r){2-3} \cmidrule(r){4-5} 
& 1\%  & 10\%  & 1\%  & 10\%  \\
\cmidrule{1-5}
% LongGAN \cite{zheng2018unsupervised} & 35.2 & 62.0 & - & -    \\
% MS$^2$L \cite{lin2020ms2l} & 33.1 & 65.2 & - & -    \\
ASSL \cite{si2020adversarial} & - & 64.3 & - & 69.8   \\
ISC \cite{thoker2021skeleton} & 35.7 & 65.9 & 38.1 & 72.5  \\
% MCC \cite{su2021modeling} & - & 60.8 & - & 65.8 \\
Colorization \cite{yang2021skeleton} & 48.3 & 71.7 & 52.5 & 78.9  \\
% CrosSCLR \cite{li20213d} & - & 67.6 & - & 73.5 \\
Hi-TRS \cite{chen2022hierarchically} & - & 70.7 & - & 74.8 \\
GL-Transformer \cite{kim2022global}) & - & 68.6 & - & 74.9 \\
HaLP \cite{shah2023halp} & 46.6 & 72.6 & 48.7 & 77.1 \\
CMD \cite{mao2022cmd} & 50.6 & 75.4 & 53.0 & 80.2 \\
CPM \cite{zhang2022contrastive} & 56.7 & 73.0 & 57.5 & 77.1 \\
HYSP \cite{franco2023hyperbolic} & - & 76.2 & - & 80.4  \\
PCM$^3$ \cite{zhang2023prompted}) & 53.8 & 77.7 & 53.1 & 82.8 \\
HiCo \cite{dong2023hierarchical} & 54.4 & 73.0 & 54.8 & 78.3 \\
UmURL \cite{sun2023unified} & 58.1 & - & 58.3 & - & \\
% RMMD\cite{HE2024127495} & 58.3 & 77.1 & 56.0 & 81.0 & \\
USDRL \cite{weng2024usdrl} & 57.3 & 80.2 & 60.7 & 84.0 \\
\cmidrule{1-5}
SkeletonMAE \cite{wu2023skeletonmae} & 54.4 & 80.6 & 54.6 & 83.5 \\
MAMP \cite{mao2023masked} & 66.0 & 88.0 & 68.7 & 91.5 \\
S-JEPA \cite{abdelfattah2024s} & 67.5 & 88.4 & 69.1 & 91.4 \\
\rowcolor{lightcyan}
GFP (Ours) & 71.8 & 88.7 & 72.9 & 92.1 \\
\bottomrule
\end{tabular}
 }
\vspace{-0.4cm}
\end{table}

%% file: 6_conclusion.tex
\section{Conclusion}
In this paper, we propose an efficient general feature prediction framework for masked skeleton modeling. Unlike conventional approaches that reconstruct low-level targets, our method employs hierarchical high-level semantic features as learning targets, significantly reducing decoder complexity. The framework integrates a target generation network that provides online learning objectives.  
We further introduce the information maximization constraints, enabling non-trivial co-optimization between networks. This mechanism simultaneously preserves local motion details and global semantics. Extensive experiments across three datasets demonstrate the effectiveness of our method, achieving state-of-the-art performance with significant training speedup over prior masked skeleton modeling approaches. 